\documentclass[lettersize,journal]{cas-dc}
\usepackage{amsmath,amsfonts}
\usepackage{algorithmic}
\usepackage{algorithm}
\usepackage{array}
\usepackage[caption=false,font=normalsize,labelfont=sf,textfont=sf]{subfig}
\usepackage{textcomp}
\usepackage{stfloats}
\usepackage{verbatim}
\usepackage{graphicx}
\usepackage{cite}
\usepackage{overpic}
\usepackage{amsfonts}
\usepackage{multirow}
\usepackage{amssymb}
\usepackage{float}
\usepackage{bm}
\usepackage{hyperref}
\usepackage{color}
\usepackage{colortbl}
\usepackage{booktabs}
\usepackage[numbers,sort&compress]{natbib}

\definecolor{mygray}{gray}{.9}

\begin{document}
\let\printorcid\relax
\title[mode = title]{PASTS: Progress-Aware Spatio-Temporal Transformer Speaker For Vision-and-Language Navigation}

\author[1]{Liuyi Wang}[style=chinese]
\author[1,2]{Chengju Liu}[style=chinese]
\cormark[1]
\author[1]{Zongtao He}[style=chinese]
\author[1]{Shu Li}[style=chinese]
\author[1]{Qingqing Yan}[style=chinese]
\author[3]{Huiyi Chen}[style=chinese]
\author[1]{Qijun Chen}[style=chinese]
\cormark[1]

\affiliation[1]{organization={Department of Control Science and Engineering, Tongji University},
    city={Shanghai},
    % postcode={201800},
    country={China}
}
\affiliation[2]{organization={Tongji Artificial Intelligence (Suzhou) Research Institute},
    city={Suzhou},
    % postcode={201800},
    country={China}
}
\affiliation[3]{organization={Rutgers University},
    city={New Brunswic},
    % postcode={201800},
    country={USA}
}

\cortext[cor1]{Corresponding author}
% \cortext[cor2]{Principal corresponding author}

\markboth{Journal of \LaTeX\ Class Files,~Vol.~14, No.~8, August~2021}%
{Shell \MakeLowercase{\textit{et al.}}: A Sample Article Using IEEEtran.cls for IEEE Journals}

% =============
% Abstract
% =============
\begin{abstract}
Vision-and-language navigation (VLN) is a crucial but challenging cross-modal navigation task. One powerful technique to enhance the generalization performance in VLN is the use of an independent speaker model to provide pseudo instructions for data augmentation. However, current speaker models based on Long-Short Term Memory (LSTM) lack the ability to attend to features relevant at different locations and time steps. To address this, we propose a novel progress-aware spatio-temporal transformer speaker (PASTS) model that uses the transformer as the core of the network. PASTS uses a spatio-temporal encoder to fuse panoramic representations and encode intermediate connections through steps. Besides, to avoid the misalignment problem that could result in incorrect supervision, a speaker progress monitor (SPM) is proposed to enable the model to estimate the progress of instruction generation and facilitate more fine-grained caption results. Additionally, a multifeature dropout (MFD) strategy is introduced to alleviate overfitting. The proposed PASTS is flexible to be combined with existing VLN models. The experimental results demonstrate that PASTS outperforms all existing speaker models and successfully improves the performance of previous VLN models, achieving state-of-the-art performance on the standard Room-to-Room (R2R) dataset.
\end{abstract}

% =============
% Keywords
% =============

\maketitle

% =============
% Introduction
% =============
\section{Introduction}
\label{sec_introduction}
The vision-and-language navigation (VLN) \cite{anderson2018vision} task, which aims to require an embodied agent to explore unstructured environments and navigate to the target location based on the fine-grained natural language instructions, has gained significant research attention owing to its flexibility and utility in various real-world scenarios. To achieve more accurate and effective navigation by capturing the relationship between visual and linguistic inputs, deep learning approaches~\cite{wang2020vision,chen2021history,chen2022think,li2022envedit,hong2021vln,guhur2021airbert,rostami2022novel, Rostami2022AnEE,dang2022unbiased,he2023mlanet, wang2023res} have been extensively explored in VLN.

\begin{figure}[tb]
\centering
\includegraphics[scale=.54]{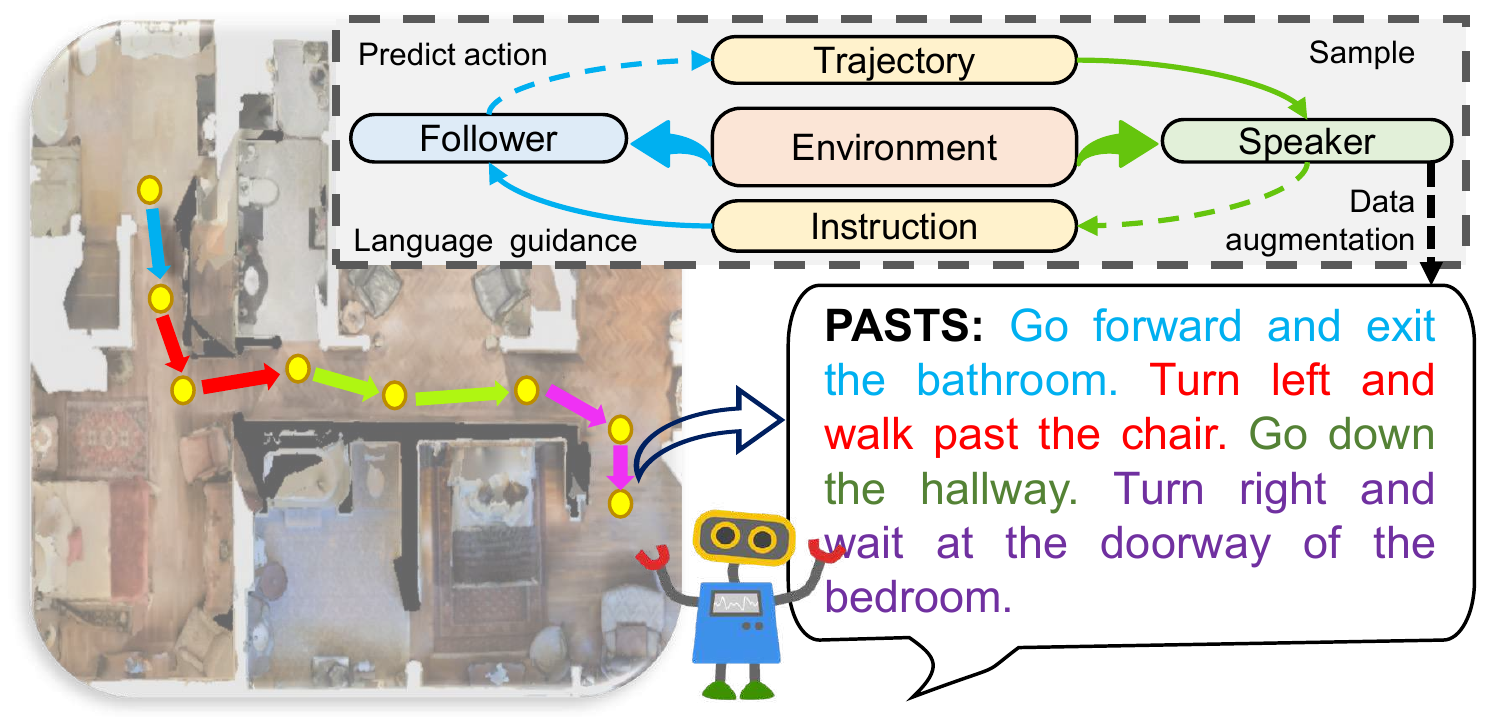}
\caption{Illustration of the follower-speaker system in VLN, where the follower aims to predict the action based on instructions, and the speaker aims to generate instructions based on trajectories. In this paper, the proposed PASTS speaker has the ability to recognize different stages for navigation (shown in different colors) and generate more accurate and fine-grained instructions.}
\label{fig_introduction}
\end{figure}

Despite the significant achievements of existing approaches in VLN, the problem of robustness becomes more obvious when the number of model parameters increases, due to the limited size of the Room-to-Room (R2R) benchmark dataset resulting from the high cost of labeling~\cite{zhang2021diagnosing}. To solve this issue, a follower-speaker system~\cite{fried2018speaker} is proposed as shown in Fig.~\ref{fig_introduction}. The follower aims to navigate toward the target location while the speaker generates descriptions of the trajectories. This approach is particularly useful when working with unlabeled trajectories which can be easily sampled in the simulator~\cite{hao2020towards}. Despite the widespread use of long short-term memory (LSTM)-based speakers~\cite{fried2018speaker,tan2019learning} for data augmentation by numerous VLN works~\cite{tan2019learning,hong2021vln,chen2021history,chen2022think,wang2022counterfactual}, the quality of the speaker has been largely overlooked. Notably, the quality of the speaker is crucial in providing pseudo-labels for the follower. A poorly-performing speaker can introduce significant noise and incorrect supervision in the VLN system. For example, if the speaker generates "turn left" or simply "turn right" instead of "turn right after leaving the bedroom" in the instructions, it can cause confusion for the follower and degrade the overall data augmentation performance of the VLN system. Therefore, this paper aims to construct a robust and capable speaker to enhance the quality of data augmentation for VLN. To achieve this goal, two factors are mainly considered as follows.
%===========
% Fig_introduction
%===========

Firstly, current speaker models lack the effective leverage of inherent long-range dependencies among the given trajectories with panorama features. From a computational perspective, the choice of spatial and temporal encoding techniques can greatly impact the performance of the proposed speaker model. In previous works, bidirectional LSTM (BLSTM)~\cite{fried2018speaker,tan2019learning}, multi-layer perception (MLP)~\cite{magassouba2021crossmap}, and hidden outputs from the follower~\cite{wu2021improved} have been used to encode the spatial features. However, BLSTM is computationally expensive and meanwhile struggles with capturing long-term dependencies. On the other hand, MLP is a simpler neural network with a lower computational cost, but it may not adequately encode multiple relationships and nuances. Using the hidden output from the follower is a cost-saving option, but it may result in the loss of key area characteristics. Regarding temporal encoding, the previous speakers~\cite{fried2018speaker,tan2019learning,wang2022counterfactual} mainly adopt the LSTM-based network to encode visual features and decode instructions. Since the recurrent structure suffers from the nature of the long-term forgetting problem, it weakens the guidance function of visual and linguistic context both in the spatial and temporal domains. Secondly, the accurate alignment of predicted instructions and the provided trajectories is a crucial aspect for the speaker. Without proper alignment, the generated instructions may lack information regarding key actions or contain repeated phrases, ultimately leading to a different path from the ground truth. To avoid this, it is necessary for the speaker to be able to segment the path targets into distinct stages. Unfortunately, current speaker models have not fully considered this aspect, creating ambiguity as to whether the generated language fully represents the navigation status.

To tackle the above problems, a novel progress-aware spatio-temporal transformer speaker (PASTS) is proposed in this work. In terms of the first limitation, a spatio-temporal transformer structure is suggested to better leverage the sequenced multiple vision and action features. Concretely, a spatial encoder is first used by employing the cross-modal multi-head attention mechanism~\cite{vaswani2017attention} to capture the correlations between 36 panoramic image patches and the oriented action-aware image. Then a temporal transformer encoder is adapted to process the successive nodes. This allows the speaker to effectively fuse observations and actions over the spatial- and temporal-dimensions. Additionally, it is imperative to enhance the practicality of the model in unfamiliar settings. However, training transformer-based architectures raises concerns about potential overfitting~\cite{wu2021unidrop} due to its fully connected nature, especially in the presence of a limited dataset. To address this issue, a multifeature dropout (MFD) strategy is introduced during training, which significantly improves the robustness of PASTS in unforeseen environments and reduces the risk of overfitting without any extra cost.

The second restriction involves an alignment issue where the speaker model may fail to properly align sub-instructions with corresponding path segments. Humans can easily split long navigation steps into smaller segments and generate descriptive phrases for each stage, ensuring proper alignment. However, it is challenging for the model to learn this skill on its own. Therefore, a speaker progress monitor (SPM) is proposed to enable the progress representation and prediction within the encoder-decoder framework. The SPM operates as an independent auxiliary task, running in parallel with the initial word prediction head. By providing additional sub-instruction and associated sub-step supervision signals, the PASTS can better identify the progress of word generation and thus increase the alignment of results.

The effectiveness of PASTS is demonstrated on the widely-used R2R dataset. Experimental results show that PASTS outperforms existing speaker models~\cite{fried2018speaker,tan2019learning,agarwal2019visual,wu2021improved,magassouba2021crossmap} and substantially improves the performance of 4 VLN follower models~\cite{tan2019learning,hong2021vln,chen2021history,chen2022think} when PASTS is applied using the back translation method. As a result, the proposed method achieves state-of-the-art VLN performance, indicating its superior performance and strong generalization ability compared to other approaches. Overall, the major contributions of this paper are summarized as follows:

1) A spatio-temporal transformer encoder is proposed to fully leverage the long-term sequenced vision features in the navigation paths, which is supposed to improve the fusion of the input observations.

2) A speaker progress monitor with a joint loss function is designed to provide strong supervision signals allowing the model to estimate its progress in instruction generation, thus facilitating more fine-grained caption results. 

3) A multifeature dropout strategy is introduced to alleviate the serious overfitting caused by the small dataset and improve the model's capability of generalization.

4) The progress-aware spatio-temporal transformer speaker can be combined with existing navigation agent methods flexibly. Adequate experiments validate the effectiveness of the proposed modules and show that PASTS can obtain state-of-the-art performance for both speaker and follower models on the VLN task.
% =============
% Fig_Overview
% =============
\begin{figure*}[!tb]
\centering
\includegraphics[scale=.56]{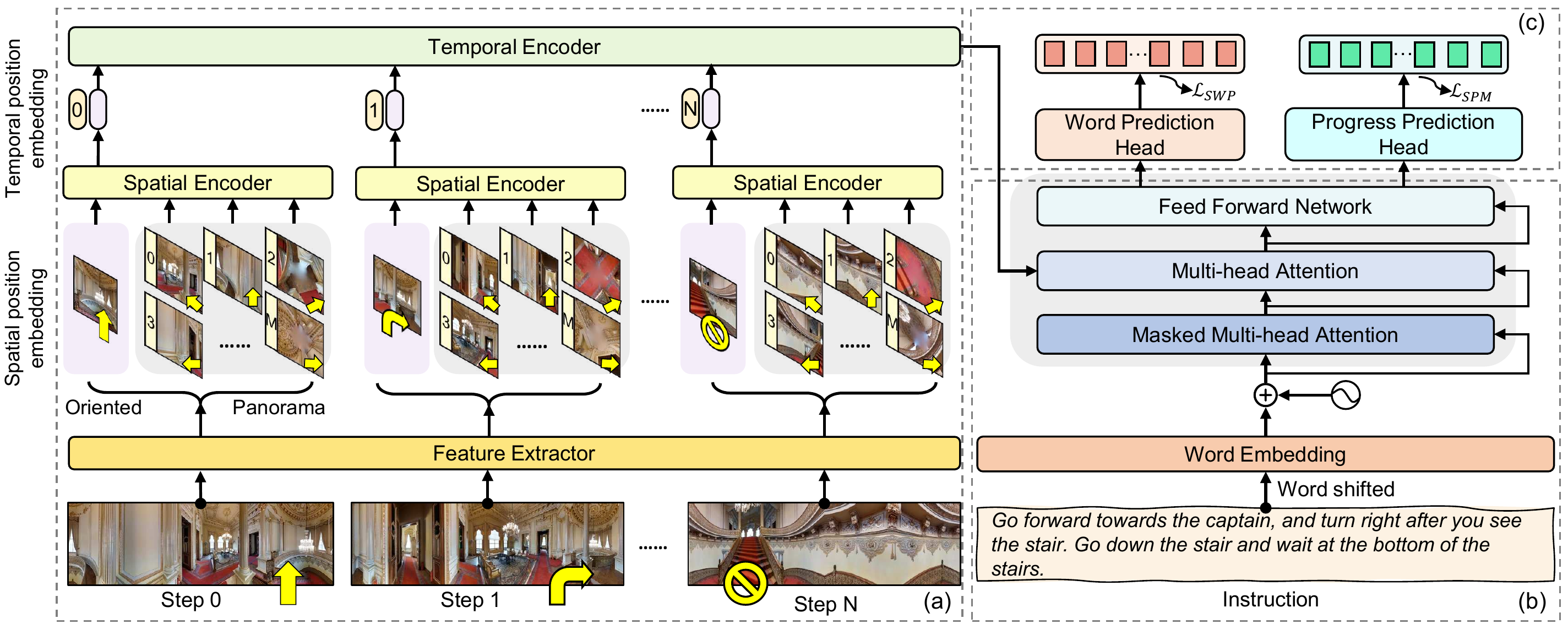}
\caption{Overall architecture of the PASTS framework, which consists of three sub-modules: (a) the spatio-temporal encoder is to integrate dominant features from environments at each location, and encode successive action features in the time dimension; (b) the generation decoder is responsible for converting the inputs of fused visual information and shifted word into a sequence of target probabilities; (c) the word prediction head and the progress prediction head are designed to predict instruction words and progress values, respectively.}
\label{fig_modeloverview}
\end{figure*}

% =============
% Related Work
% =============
\section{Related Work}
\label{sec_relatedwork}

\subsection{Vision-and-Language Navigation}
\label{subsec_vln}
VLN is first proposed by Anderson et al. \cite{anderson2018vision} based on Matterport3D \cite{chang2017matterport3d}, a large-scale reinforcement learning environment based on real imagery, and has been a focus of significant research interest in recent years \cite{zhangSurveyVisualNavigation2022,zhang2020language,wang2020vision,an2021neighbor,chen2021history,dang2022search,wang2023dual}. A wide range of strategies have been proposed to enhance navigation capability from diverse perspectives. For instance, reinforcement learning techniques \cite{wang2019reinforced} have been introduced to improve the decision-making process of VLN models, and the fusion of multiple modal input features \cite{chen2022reinforced,sun2021depth} has been explored to augment the perceptual capacity. Additionally, constructing historical memory graph~\cite{hong2021vln,chen2021history,chen2022think} and incorporating auxiliary tasks \cite{zhu2020vision,ma2019self,zhao2022target} have been considered to enhance the capability of history dependency and to enrich the inference modes. Despite significant progress, overfitting is still a critical challenge in VLN. The limited scale of the dataset can lead to poor performance of the agent in unseen environments. Therefore, some data augmentation and dataset expansion approaches are proposed to improve the generalization capability ~\cite{agarwal2019visual,fu2020counterfactual,lin2021adversarial,liu2021vision,li2022envedit,hao2020towards,guhur2021airbert}. 

The speaker model \cite{fried2018speaker,agarwal2019visual,tan2019learning} has been proven to be a valuable tool for data augmentation in VLN. Concretely, Speaker-Follower \cite{fried2018speaker} first proposes to use an independent model to automatically generate pseudo labels for data augmentation. WIP-Speaker \cite{agarwal2019visual} is a work-in-progress speaker model that adopts hard attention in two stages. Env-Speaker \cite{tan2019learning} uses the BLSTM and soft-attention mechanism to extract trajectories and introduces the back translation method into VLN. Zhao et al.~\cite{zhao2021evaluation} study the evaluation indicators for evaluating the performance of a speaker model. Wang et al. \cite{wang2022counterfactual} and Dou et al. \cite{dou2022foam} both suggest training the speaker and the follower jointly. All of the above speaker models are based on the LSTM architecture. Some methods like Cmp-Speaker \cite{magassouba2021crossmap} and Imp-Speaker \cite{wu2021improved} attempt to build transformer-based speaker and follower models, but their performance improvements are limited. In summary, prior studies have demonstrated the validity of using an independent instruction generator to augment the VLN dataset and improve model performance. However, these approaches have been limited in their ability to effectively leverage panoramic images across different steps and ensure the alignment of the predicted instructions and the sampled paths. As a result, these limitations pose potential risks to the efficacy of speaker-follower systems. Therefore, this work proposes a spatio-temporal transformer speaker to generate more fine-grained instructions of higher accuracy.

\subsection{Transformers for Visual Captioning}
Models for visual captioning tasks (such as image captioning~\cite{zha2019context}, video captioning~\cite{STAT2020yan}, and visual question answering~\cite{self2021zhong}) usually follow an encoder-decoder framework and can be regarded as performing a task of neural machine translation from image to text. Typically, convolutional neural network (CNN) -- recurrent neural network (RNN) architectures are used to encode the images as feature vectors and then decode these vectors into sentences in a recurrent manner~\cite{fried2018speaker,deep2019xiao,zha2019context}. Recently, the transformer~\cite{zhou2016attention} and its extensions~\cite{devlin2018bert,brown2020language} have shown remarkable improvements in various tasks, leading to the gradual replacement of the RNN architecture~\cite{zhou2018end,guo2020normalized,luo2021dual}. Motivated by this progress, the transformer-based architecture is employed as the backbone of the proposed speaker model. However, while the speaker in VLN and visual captioning tasks share some similarities, it is important to note that they differ in certain aspects related to input form and field of application. For instance, visual captioning methods are typically designed to handle single-oriented images and may not be capable of capturing the panoramic surroundings that are crucial to VLN. Moreover, VLN involves varying actions at each step, making it a more challenging task for the speaker model. Despite drawing inspiration from visual captioning methods, developing a stronger speaker model for VLN remains an open research problem.

\subsection{Auxiliary Tasks}
\label{subsec_auxiliary}
In the field of machine learning, auxiliary tasks have been widely employed to improve data efficiency and robustness~\cite{veeriah2019discovery,gu2019scene,trinh2018learning}. Existing methods have demonstrated that the performance of VLN models can also be boosted with the assistance of auxiliary tasks. Progress Monitor~\cite{ma2019self} aims to ensure that the grounded instruction correctly reflects the navigation progress. Huang et al.~\cite{huang2019transferable} introduced two auxiliary tasks, cross-modal alignment (CMA) and next visual scene (NVS), which involve assessing the fit between a given instruction--path pair and predicting latent representations of future visual inputs, respectively. AuxRN~\cite{zhu2020vision} is a framework that includes four self-supervised auxiliary reasoning tasks to take advantage of additional training signals derived from semantic information. Considering that an auxiliary task can effectively introduce additional strong supervision signals by prior knowledge and the inherent physical characteristics of the specific task at hand, a novel SPM is designed to force the model to learn to recognize its generation progress during training and thus improve the alignment between trajectories and instructions.

% =============
% Proposed Methods
% =============
\section{Method}
\label{sec_method}
\textbf{Problem Setup}
% In this paper, we mainly focus on researching a TIG model and verifying the effectiveness of our model by combining it with VLN models.
Matterport3D~\cite{chang2017matterport3d} is used as the simulator to produce the connected graph $G=\{P,\xi \}$, where $P$ represents navigable viewpoints and $\xi$ represents the connections between these viewpoints. In the R2R dataset~\cite{anderson2018vision}, the data is annotated as pairs of trajectory $\tau=\{p_1,p_2,...,p_N\}$ and instruction $I=\{w_1,w_2,...,w_L\}$ where $p_i$ and $w_i$ represent the visited nodes and words, and $N$ and $L$ denote the length of the path and the instruction, respectively. At each step, the agent can observe a visual environmental panorama, which includes 3 perspectives that each has 12 images of $30^\circ$. The resolution of each image is $640\times 480$. Let $d_v$ and $d_o$ represent the dimension of image features and orientation features. Specifically, the actual forward path locates at one of the images at each step, and this set of images with their offset orientations is assigned as the action-level features $A=\{V^a;\gamma^a \}$, where $V^a=\{v_1^a,v_2^a,...,v_N^a\} \in \mathbb{R}^{N \times d_v}$ denotes $N$ image features with the angle set $\gamma^a=\{\gamma^a_1,\gamma^a_2,...,\gamma^a_N\}\in \mathbb{R}^{N \times d_o}$. Similarly, the environment-level features $E=\{V^e;\gamma^e\}$ is composed of the panoramic image set $V^e = \{[v_{p,i}^e]_{i=1}^{M}\}_{p=1}^N \in \mathbb{R}^{N \times M \times d_v}$ and the angle set $\gamma^e=\{[\gamma_{p,i}^e]_{i=1}^{M}\}_{p=1}^N \in \mathbb{R}^{N \times M \times d_o}$, where $M$ represents the number of image per panorama view. 

The goal of the VLN follower is to enable an agent to find a correct navigation trajectory in a real indoor environment with the assistance of instructions and visual observations. In contrast, the speaker aims to predict the probability of a set of words in the instructions for a given trajectory with visual observations. Formally, the probability of action prediction and word prediction can be written as Eq.~\eqref{equation_SWP} and \eqref{equation_SWP2}, respectively.
\begin{align}
\label{equation_SWP}
    P_v(p_1,...,p_N | I,E) &= \prod_{i=1}^N P_v(p_i|p_1,...,p_{i-1},I,E) \\ \label{equation_SWP2}
    P_t(w_1,...,w_L | A,E) &= \prod_{l=1}^L P_t(w_l|w_1,...,w_{l-1},A,E) 
\end{align}

% =============
% PASTS
% =============
\subsection{Spatio-Temporal Transformer Encoder}
\label{subsec_transpeaker}
\begin{figure}[tbp]
    \centering
    \includegraphics[scale=0.58]{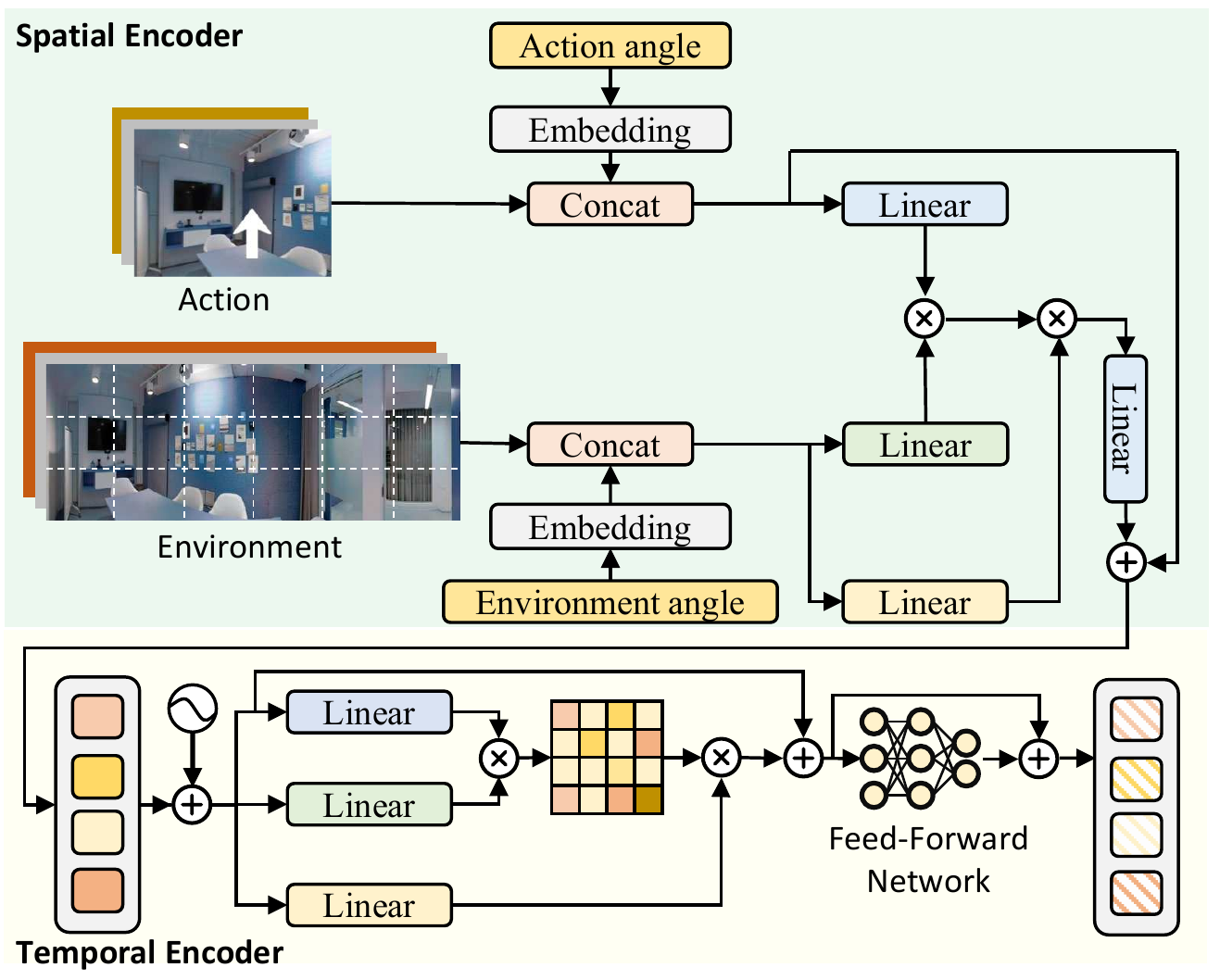}
\caption{Illustration of the spatial encoder (top) and the temporal encoder (bottom). The spatial encoder is used to effectively fuse the action embedding and the environment embedding in each step, and the temporal encoder is applied to capture the internal connection between different navigation steps.}
    \label{fig_aecm}
\end{figure}

As shown in Fig.~\ref{fig_modeloverview}, the core of PASTS is constructed based on a sequence-to-sequence transformer architecture to explore a powerful network for instruction generation. To better use the action and environment information throughout time and space, a novel spatio-temporal transformer encoder structure is first designed based on the cross-modal attention mechanism module, which is illustrated in Fig.~\ref{fig_aecm}.

\textbf{The spatial encoder} The inputs to the speaker consist of two types of observations: a set of actions $A\in \mathbb{R}^{N \times d_p}$ and a set of environmental observations $E\in \mathbb{R}^{N \times M \times d_p}$. Specifically, every panorama consists of 36 image patches for each viewpoint. The reasonable fusion of visual features is critical and influences the model's cognitive level. Previous most commonly used methods~\cite{fried2018speaker,tan2019learning} have used BLSTM with a soft attention mechanism~\cite{zhou2016attention} to capture the most important information in panoramic sequences. In contrast, considering the complexity of BLSTM networks and the difficulty of proper training~\cite{pascanu2013difficulty}, we first propose a spatial encoder structure to better combine the features of actions with those of environmental observations, enabling the model to learn to focus on more relevant parts of the input views. 
% =============
% Fig_spatio-temporal encoder
% =============
Since the action features guide the main variants among paths, $A$ is regarded to be the query and $E$ is regarded to be the key and value. It is beneficial to linearly project the queries, keys and values through different linear transformations by using a multihead attention mechanism, allowing the model to focus on different representation subspaces at different locations. After the concatenation of the results of the multiple attention heads, a residual connection~\cite{he2016deep} followed by layer normalization~\cite{ba2016layer} is applied to achieve the result $Z \in \mathbb{R}^{N \times d_m}$. To increase the generalization ability, a few additional feature dropouts are included in the model, which will be further discussed in Sec.~\ref{subsec_EFD}. Afterwards, the formulas for the spatial encoder are given in Eq.~\eqref{equation_AECM}--\eqref{equation_AECM_2}.
\begin{align} \label{equation_AECM}
\widetilde{A} &= (\text{Dropout}(V^a);\gamma^a)W_a \\
\widetilde{E} &= (\text{Dropout}(V^e);\gamma^e)W_e \\
Q^i_{\widetilde{A}} = \widetilde{A}W_q^i,\,K^i_{\widetilde{E}} &= \widetilde{E}W_k^i,\,V_{\widetilde{E}}^i = \widetilde{E}W_v^i \\
\text{head}_i &= \text{softmax}(\frac{Q_{\widetilde{A}}^i {K_{\widetilde{E}}^{^i}}^T}{\sqrt{d_k}})V_{\widetilde{E}}^i \\
\text{MultiHead}(Q,K,V) &= \text{Concat}(\text{head}_1,...,\text{head}_H)W_o \\ \label{equation_AECM_2}
Z &= \text{LayerNorm}(\text{MultiHead}+\widetilde{A})
\end{align}
where $W_a \in \mathbb{R}^{d_p \times d_m}$, $W_e \in \mathbb{R}^{d_p \times d_{m}}$, $W_q^i\in \mathbb{R}^{ d_{m}\times d_q}$, $W_k^i \in \mathbb{R}^{d_{m} \times d_k}$, $W_v^i \in \mathbb{R}^{d_m\times d_v}$, $W_o\in \mathbb{R}^{Hd_v\times d_m}$. After the aggregation of action and environment features, the output of the spatial encoder is $Z \in \mathbb{R}^{N \times d_m}$. The multihead attention mechanism boosts the model's capability to capture the important semantics of the surroundings and fuse them into the oriented image features, reducing much useless noise and interference.

\textbf{The temporal encoder} After fusing different spatial information at each time stage via the spatial encoder, a temporal encoder is employed to enable the model to learn the inherent correlations between various time stages. This is crucial for the speaker model since the desired instruction is specified with respect to the actual navigation forward process. Concretely, the spatial-fused features $Z=\{z_1,z_2,...,z_N\}$ among different steps are then input into the temporal encoder with $L$ layers as the independent tokens. Position embeddings $PE(\cdot)$~\cite{vaswani2017attention} are added to the $Z$ to retain the positional information. Each encoder layer consists of a multihead self-attention (MSA) layer followed by a small feedforward neural network (FFN). Because the generated trajectory instructions are strongly related to the actual order of navigation, the position encoding is incorporated into the visual sequence embeddings. Similar to the spatial encoder, the residual connections are employed around each of the sub-layers, followed by layer normalization. Let $l$ denote the $l$-th layer, where $l=1...L$, the formula of the temporal encoder is as Eq.~\eqref{equation_temporal_encoder}--\eqref{equation_temporal_encoder_2}:
\begin{align} \label{equation_temporal_encoder}
Z_0 &= PE(Z) \\
Z'_l &= \text{LayerNorm}(\text{MSA}(Z_{l-1})+Z_{l-1}) \\
Z_l &= \text{LayerNorm}(\text{FFN}(Z'_l)+Z'_l) \label{equation_temporal_encoder_2}
\end{align}

\textbf{The text decoder} The decoder part follows the transformer decoder architecture. Since language generation is an autoregressive process, it is necessary to ensure that each predicted word depends only on the previous ones. Therefore, the word embeddings are offset by a special token \textless BOS\textgreater, and a mask function is applied to the attention matrix to mask out illegal positions. Position encoding is also added to the word embeddings to capture the relative positions of the tokens in the sequence. Supposing that the target vocabulary is $d_{\text{vocab}}$ and that each instruction contains a maximum of $L$ words, a linear layer and a softmax operation are applied to convert the values in the last hidden layer of the decoder, $g_{t,d} \in \mathbb{R}^{L\times d_m}$, into projected probabilities, $p_{t,d} \in \mathbb{R}^{L \times d_{\text{vocab}}}$. For symmetry, we call the head that performs this task the speaker word projector (SWP), which is optimized with a cross-entropy loss, as shown in Eq.~\eqref{euqation_loss_SWP}:
{\setlength\abovedisplayskip{0pt}
\setlength\belowdisplayskip{1pt}
\begin{equation}
\label{euqation_loss_SWP}
\mathcal{L}_{\text{SWP}} = -\sum_{i=1}^L \log (f_\theta(w_i^*|w_{1:i-1}^*,A,E))
\end{equation}}
where $\theta$ represents the parameters of our model, $w^*_{1:l}$ is the target ground-truth sequence, and $f_\theta(\cdot)$ denotes the predicted probability that the target word is in the $i$-th location as calculated through the model with actions $A$ and environmental observations $E$.

% =============
% SPM
% =============
\subsection{Speaker Progress Monitor (SPM)}
\label{sub_SPM}
Previous speaker models have overlooked the fundamental correlation embedded within the given trajectory and thus fail to relate each word to the progress along the given path. This might lead to misalignment between the trajectories and instructions. Therefore, an SPM module is proposed to enable the model to estimate the progress of instruction generation, which serves as an auxiliary task during training. 
% =============
% Fig_SPM
% =============
\begin{figure}[tbp]
\centering
\includegraphics[scale=.5]{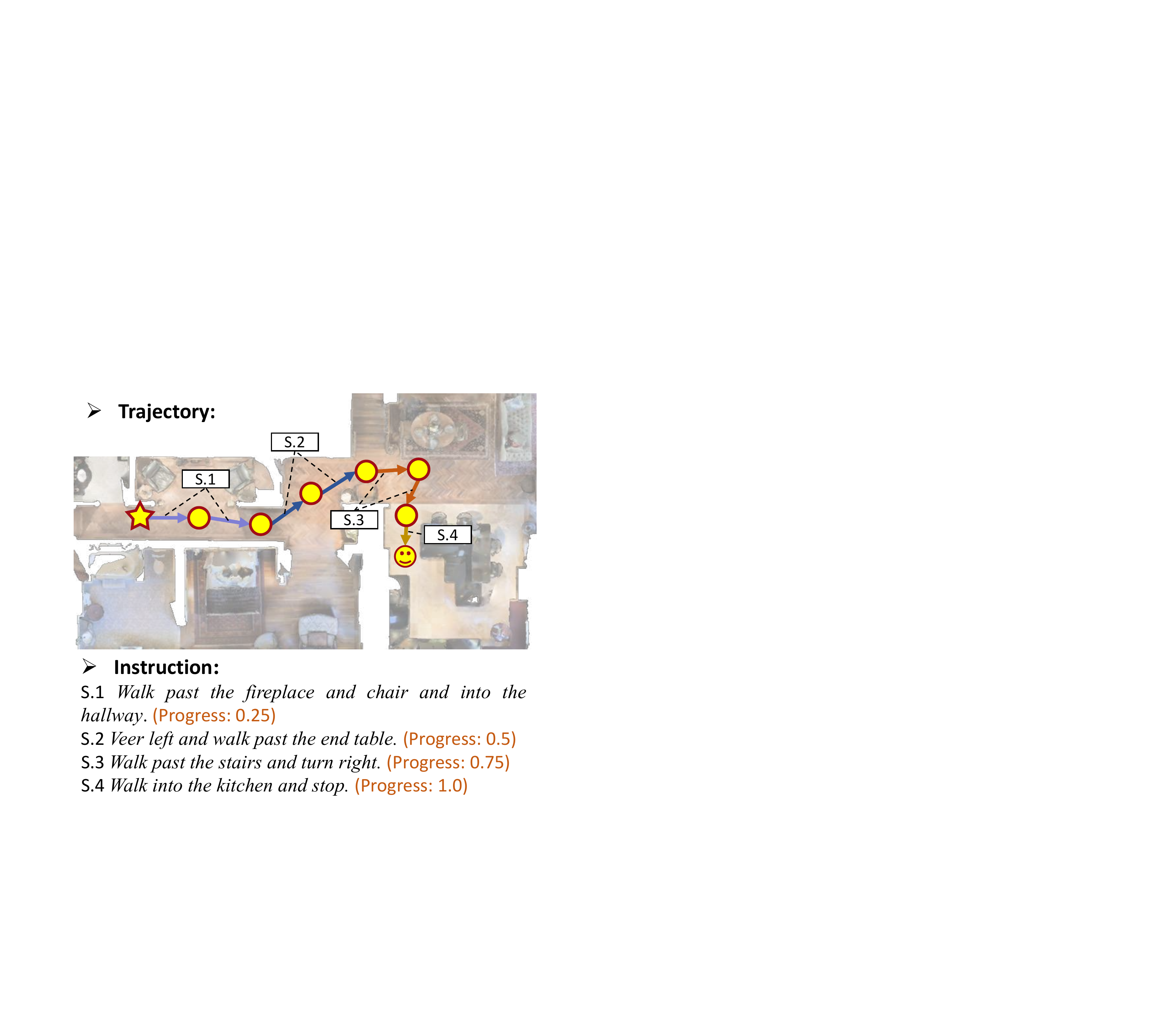}
\caption{Illustration of the speaker progress monitor (SPM). The complete trajectories and instructions are divided into several subsets. Each word is assigned the corresponding progress value (shown in brown) to align the instructions and trajectories. Different colors are used to represent different navigation stages.} 
\label{fig_fg}
\end{figure}
To ensure that each instruction word corresponds to a specific stage of the input trajectory, the FGR2R dataset~\cite{hong2020sub}, which contains the fine-grained annotations for sub-paths and sub-instructions, is adapted to provide the ground-truth progress signal for each word. Formally, the data pair provided by R2R is $\{(\tau,I)\}$, where $\tau=\{p_1,p_2,...,p_N\}$ and $I=\{w_1,w_2,...,w_L\}$ represent the trajectory with $N$ nodes and the instruction with $L$ words, respectively. By dividing the long trajectory into $K$ subsets, the data pair in FGR2R is in the format $\{ (\tau'_1,I'_1), (\tau'_2,I'_2),...,(\tau'_K, I'_K) \}$, where each subset represents a smaller navigation task. Fig.~\ref{fig_fg} shows an example that a whole trajectory can be divided into four navigation stages. Therefore, the individual word can be assigned the corresponding progress values. Suppose that $M_{progress}=\{m_1,m_2,...,m_L\}$ represents the instruction progress set, and the progress value of each word follows Eq.~\eqref{equation_SPM_1}:
\begin{align} 
    \label{equation_SPM_1}
    % m_{i\,(w_{i} \in I_{j}^{'})} = \frac{\text{Index}(I_j')}{k}
    m_{i\,(w_{i} \in I_{j}^{'})} = \frac{j}{K}
\end{align} 

With the above definition, all words belonging to the same subset are associated with the same progress signal. To supervise the model with the ground progress value, the SPM is integrated into the model as an auxiliary task. Concretely, in parallel with the SWP, the SPM module is employed after the last hidden layer of the decoder. The SPM contains two linear transition layers, a rectified linear unit (ReLU) activation layer and a dropout layer, as expressed in Eq.~\eqref{equation_SPM_2}--\eqref{equation_SPM_3}: 
\begin{align} \label{equation_SPM_2}
    Z'_{l,pm} &= \text{ReLU}(h_{l,d}W_s + b_s) \\ \label{equation_SPM_3}
    Z_{l,pm} &= \text{Dropout}(Z'_{l,pm})W_p + b_p
\end{align}
where $W_s \in \mathbb{R}^{d_m\times d_{\text{hidden}}}$, $b_s \in \mathbb{R}^{L \times d_{\text{hidden}}}$, $W_p \in \mathbb{R}^{d_{\text{hidden}}\times 1}$, and $b_p \in \mathbb{R}^{L \times 1}$ are the learned parameters. In this work, we use $d_m=256$ and $d_{\text{hidden}}=512$. Since SPM is a regression model, the mean squared error (MSE) loss shown in Eq.~\eqref{equation_SPM4} is applied as the loss function:
\begin{align} \label{equation_SPM4}
    \mathcal{L}_{\text{SPM}} &= -\frac{1}{2L}\sum_{l=1}^L (m_l^*-Z_{l,pm})^2
\end{align}
where $m_l^*$ represents the ground-truth progress value. Finally, the SWP and the SPM are jointly trained in an end-to-end manner as Eq.~\eqref{equation_SPM5}:
% \belowdisplayshortskip
\begin{align} \label{equation_SPM5}
    \mathcal{L}_{total} = \lambda\mathcal{L}_{\text{SWP}} + (1-\lambda)\omega\mathcal{L}_{\text{SPM}}
\end{align}
where $\lambda$ is a weight used to control the balance between the two losses and $\omega$ is used to ensure that the two losses are of the same order of magnitude.

% =============
% MFD
% =============
\subsection{Multifeature Dropout (MFD)}
\label{subsec_EFD}
\begin{figure}[tbp]
\centering
\setlength{\abovecaptionskip}{0.cm}
\includegraphics[scale=.63]{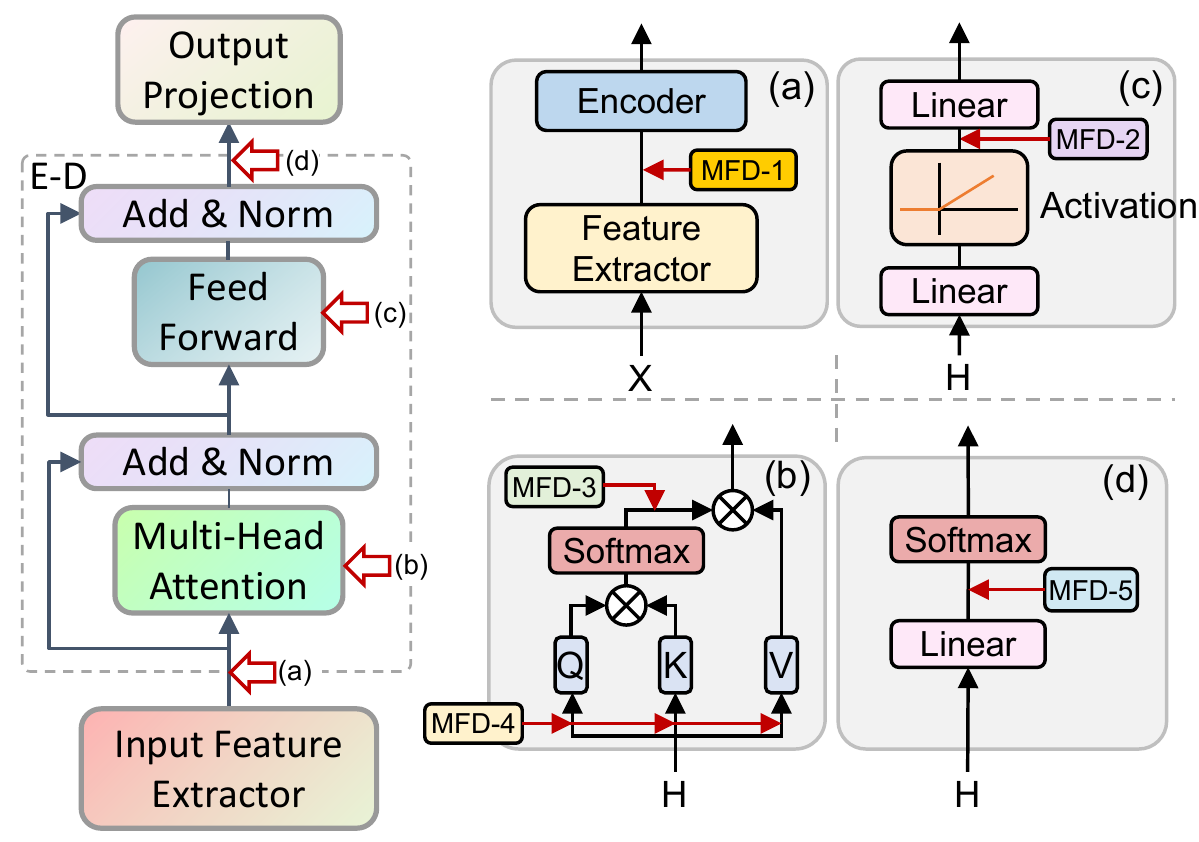}
\caption{Illustration of the multifeature dropout (MFD). In order to avoid serious overfitting, five dropouts at different positions are proposed to increase the diversity of the network structure. (a) represents the input representation, (b) indicates the attention mechanism, (c) represents the feed-forward network, and (d) is the output projection.}
\label{fig_mfd}
\end{figure}
Although the transformer architecture has shown remarkable performance in various domains, one major challenge in training such models is overfitting. While the speaker is designed to alleviate the overfitting problem for VLN, it also poses a similar problem for speaker training, particularly when the training data from R2R is limited. Inspired by the regularization properties of dropout~\cite{baldi2014dropout}, which randomly deactivates a fraction of the neurons during each training iteration, a multifeature dropout (MFD) strategy is proposed to augment the features and reduce overfitting during training.

The illustration of MFD is shown in Fig.~\ref{fig_mfd}. Concretely, the four fundamental modules of the transformer-based structure are considered: the input feature extractor, the feed-forward network, the attention mechanism, and the output projection module. First, the environment dropout \cite{tan2019learning} has been demonstrated to be extremely effective in creating various environments in VLN. Therefore, this type of dropout is applied to the input features after the feature extractor, and named MFD-1. The angles are left unchanged since every angle is essential in navigation; even minor mutations may lead to considerable ambiguity. Additionally, the basic transformer~\cite{vaswani2017attention} contains two specific types of feature dropout, which locate after the activation layer of the FFN modules and after the softmax function of the attention modules, which are called MFD-2 and MFD-3, respectively. To further increase the diversity of the structure, following UniDrop~\cite{wu2021unidrop}, two additional types of feature dropout, MFD-4 and MFD-5, which locates before the calculation of $Q,K,V$ and before the final output projection respectively, are added in the framework as well. Above all, the total five types of feature dropout applied in PASTS are summarized as follows:

1) MFD-1 (environment dropout): After using the image feature extractor to capture the visual observation, MFD-1 is adopted to randomly mask different regions of environment features.

2) MFD-2 (activation dropout): The transformer layers comprise of FFN and attention mechanism. In FFN, MFD-2 is applied after the activation function between the two linear transition layers.

3) MFD-3 (attention dropout): In the attention modules, MFD-3 is applied to the attention matrix, which is after the matrix product $QK^T$ but before weighting by the value matrix $V$.

4) MFD-4 (query, key, and value dropout): To improve the diversity of queries and keys in the attention mechanism, MFD-4 is applied after linearization with $W_q$, $W_k$, and $W_v$ and before the matrix product of $Q$ and $K$.

5) MFD-5 (output dropout): A softmax function with linear projection is used after the last hidden layer of the decoder to output the outputs. MFD-5 is applied before the linear transition layer of the final classification module.

\subsection{Training Procedures}
\label{sub_training}
 The speaker-follower system~\cite{fried2018speaker} includes two agents: the follower is to follow the instructions and navigate to the target area, and the speaker is to generate pseudo instructions for augmenting data pairs. In the previous sections, the proposed speaker PASTS and its training loss are mainly discussed. Here, the training strategy of the follower and the back translation method are briefly introduced.

\textbf{Follower training} Let $I$, $T$, $E$ denote instructions, trajectories and environments, respectively. The encoder-decoder follower is to learn the mapping of $\{I,E\} \rightarrow T$. Since this paper mainly focuses on the speaker study, the training strategy of the follower keeps consistent with the corresponding approaches. Following the operation in recent works~\cite{tan2019learning,hong2021vln,chen2021history}, a mixture of imitation learning (IL) and reinforcement learning (RL) is adopted to train the follower models. The IL relies on expert action at each step, while RL samples actions according to the policy $\pi$. Specifically, the loss of IL is used for off-policy learning and is written as
{\setlength\abovedisplayskip{0pt}
\setlength\belowdisplayskip{1pt}
\begin{equation}
\label{euqation_loss_agent_il}
\mathcal{L}_{\text{IL}} = -\sum_{n=1}^N \log \pi(\tau_i^*)
\end{equation}}

Reinforcement learning is applied for on-policy learning, where the optimization objectiveness is expressed as
{\setlength\abovedisplayskip{0pt}
\setlength\belowdisplayskip{1pt}
\begin{equation}
\label{euqation_loss_agent_rl}
\mathcal{L}_{\text{RL}} = -\sum_{n=1}^N \log \pi(\hat{\tau}_i^h)\Lambda_n
\end{equation}}
where $\Lambda_n$ indicates the advantage in the Actor-Critic RL algorithm \cite{mnih2016asynchronous}. Thus, the total loss of the follower training is $\mathcal{L}_{\text{follower}}=\mathcal{L}_{\text{IL}}+\lambda_{\text{RL}}\mathcal{L}_{\text{RL}}$, where $\lambda_{\text{RL}}$ is used to adjust the weight proportion of two losses. When training DUET~\cite{chen2022think}, RL is replaced with the pseudo interactive demonstrator (PID) strategy.

\begin{figure}[tbp]
\centering
\includegraphics[scale=.57]{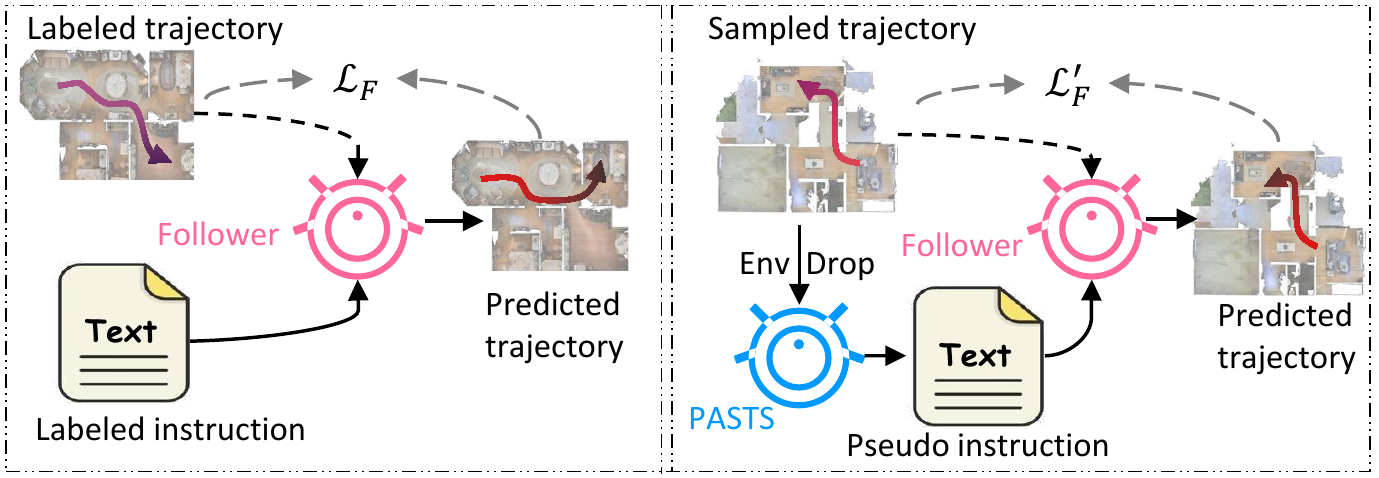}
\caption{Illustration of the training procedure of the follower via the speaker based on the back translation method.} 
\label{fig_speaker_follower}
\end{figure}

\textbf{Back translation} As shown in Fig.~\ref{fig_speaker_follower}, the central idea of back translation is to translate sampled paths $T'$ and augmented environments $E'$ into pseudo instructions $I'$ and use new tuples $\{I',E',T'\}$ to augment the training dataset. In this work, PASTS is first trained based on the optimization objective in Eq.~\eqref{equation_SPM5} in the original training dataset $\mathcal{D}$. Subsequently, with the large number of trajectories sampled by~\cite{hao2020towards}, the trained PASTS is utilized in conjunction with environment dropout to generate new data pairs $\mathcal{D'}$. For stabilizing the optimization of back translation, as~\cite{tan2019learning}, $\mathcal{D}$ and $\mathcal{D'}$ are alternated employed during training while sharing environment dropouts in the same batch. This enables the follower to be exposed to a wider range of environments, thus enhancing its generalization ability.
% ============
% Experimental setting
% ============
\begin{table}[!htb]
\caption{Experiment settings.}
\label{tab_experiment_settings}
\begin{tabular}{l|l}
\toprule
Item & Setting \\ \midrule
Layer of temporal encoder & 6 \\
Layer of decoder & 6 \\
Number of head & 6 \\
Dimension of head & 64 \\
Hidden dimension of encoder & 512 \\
Hidden dimension of decoder & 256 \\
Hidden dimension of FFN & 1024 \\
Dropout ratio of $\text{MFD}_{1}$ & 0.3 \\
Dropout ratio of $\text{MFD}_{2-5}$ & 0.2 \\
Learning rate & $5\times10^{-5}$ \\
Batch size & 64 \\
Optimizer & Adam \\
Iteration & 80,000 ($\sim$6.5 hours) \\ \bottomrule
\end{tabular}
\end{table}
% ============
% Speaker comparison
% ============
\begin{table*}[!htb]
\caption{Performance comparison with existing speaker models on the R2R dataset. All the values are reported as percentages (\%). The underlined value represents the maximum value using the same feature (ResNet-152), while the bolded value represents the maximum value using different features.}
\label{table_resultsTIG}
\centering
\resizebox{\textwidth}{!}{
\begin{tabular}{l|l|ccccc|ccccc}
\toprule
\multirow{2}{*}{Method} & \multirow{2}{*}{Feature} & \multicolumn{5}{c|}{R2R Validation-Seen} & \multicolumn{5}{c}{R2R Validation-Unseen} \\ 
 & & BLEU-1$\uparrow$ & BLEU-4$\uparrow$ & ROUGE-L$\uparrow$ & CIDEr$\uparrow$ & SPICE$\uparrow$ & BLEU-1$\uparrow$ & BLEU-4$\uparrow$ & ROUGE-L$\uparrow$ & CIDEr$\uparrow$ & SPICE$\uparrow$ \\ \midrule
Speaker-Follower~\cite{fried2018speaker} & ResNet-152 & 53.7 & 15.5 & 35.0 & 12.1 & 20.3 & 52.2 & 14.2 & 34.6 & 11.4 & 18.8 \\
WIP-Speaker~\cite{agarwal2019visual} & ResNet-152 & 54.9 & 15.7 & 35.2 & 13.7 & 21.4 & 54.8 & 15.0 & 35.3 & 13.2 & 19.7 \\
Env-Speaker~\cite{tan2019learning} & ResNet-152 & 56.8 & 18.2 & 36.4 & 18.0 & 22.6 & 55.7 & 16.7 & 35.7 & 14.4 & 20.5 \\ 
Cmp-Speaker~\cite{magassouba2021crossmap} & ResNet-152 & 53.7 & 16.2 & 34.4 & 15.3 & 21.7 & 52.5 & 14.6 & 33.9 & 11.8 & 19.5 \\ 
Imp-Speaker~\cite{wu2021improved} & ResNet-152 & 58.6 & 19.4 & 36.9 & 18.8 & 24.0 & 55.6 & 16.7 & 35.4 & 13.8 & 20.8 \\ \hline
PASTS (Ours) & ResNet-152  & \underline{59.6} & \underline{19.7} & \underline{37.8} & \underline{19.5} & \underline{25.6} & \underline{57.1} & \underline{17.6} & \underline{36.6} & \underline{15.2} & \underline{21.5} \\
PASTS (Ours) & CLIP-B/32 & 60.0 & 19.8 & 37.7 & 20.5 & 25.6 & 57.6 & \textbf{18.6} & 36.9 & 16.5 & 22.3 \\
PASTS (Ours) & ViT-B/16  & \textbf{60.9} & \textbf{20.5} & \textbf{38.2} & \textbf{21.7} & \textbf{26.3} & \textbf{57.9} & 18.5 & \textbf{37.0} & \textbf{16.8} & \textbf{22.9} \\ \hline
\end{tabular}}
\end{table*}
% ============
% Follower comparison
% ============
\begin{table*}[!htb]
\caption{Performance comparison with existing agent models on the R2R dataset. All the values are reported as percentages (\%).}
\label{table_resultsVLN}
\centering
\resizebox{\textwidth}{!}{
\begin{tabular}{l|cccc|cccc|cccc}
\toprule
\multirow{2}{*}{Method} & \multicolumn{4}{c|}{R2R Validation-Seen} & \multicolumn{4}{c|}{R2R Validation-Unseen} & \multicolumn{4}{c}{R2R Test (Unseen, Single-run)} \\
 & NE$\downarrow$ & TL & SR$\uparrow$ & SPL$\uparrow$ & NE$\downarrow$ & TL & SR$\uparrow$ & SPL$\uparrow$ & NE$\downarrow$ & TL & SR$\uparrow$ & SPL$\uparrow$ \\ \midrule
Random \cite{anderson2018vision}& 9.45 & 9.58 & 16 & - & 9.23 & 9.77 & 16 & - & 9.79 & 9.93 & 17 & 12 \\
Seq-to-Seq \cite{anderson2018vision}& 6.01 & 11.33 & 39 & - & 7.81 & 8.39 & 22 & - & 7.85 & 8.13 & 20 & 18 \\ 
Speaker-Follower \cite{fried2018speaker} & 3.36 & - & 66 & - & 6.62 & - & 35 & - & 6.62 & - & 35 & 28 \\
Self-Monitoring \cite{ma2019self}& 3.22 & - & 67 & 58 & 5.52 & - & 45 & 32 & 5.67 & - & 48 & 35 \\
RCM \cite{wang2019reinforced}& 3.53 & 10.65 & 67 & - & 6.09 & 11.46 & 43 & - & 6.12 & 11.97 & 43 & 38 \\
AuxRN \cite{trinh2018learning}& 3.33 & - & 70 & 67 & 5.28 & - & 55 & 50 & 5.15 & - & 55 & 51 \\
PREVALENT \cite{hao2020towards}& 3.67 & 10.32 & 60 & 65 & 4.73 & 10.19 & 57 & 53 & 4.75 & 10.51 & 54 & 51 \\
RelGraph \cite{hong2020language}& 3.47 & 10.13 & 67 & 65 & 4.73 & 9.99 & 57 & 53 & 4.75 & 10.29 & 55 & 52 \\
NVEM \cite{an2021neighbor}& 3.44 & 11.09 & 69 & 65 & 4.27 & 11.83 & 60 & 55 & 4.37 & 12.98 & 58 & 54 \\
HOP \cite{qiao2022hop}& 2.72 & 11.26 & 75 & 70 & 3.80 & 12.27 & 64 & 57 & 3.83 & 12.68 & 64 & 59 \\ \hline
EnvDrop \cite{tan2019learning} & 3.99 & 11.00 & 62 & 59 & 5.22 & 10.70 & 52 & 48 & 5.23 & 11.70 & 51 & 47 \\ \rowcolor{mygray}
EnvDrop+PASTS (Ours) & \textbf{3.38} & 11.92 & \textbf{69} & \textbf{65} & \textbf{4.80} & 15.26 & \textbf{55} & \textbf{50} & 5.30 & 10.90 & \textbf{56} & \textbf{52} \\ \hline
RecBERT \cite{hong2021vln}& 2.82 & 11.12 & 74 & 69 & 4.24 & 12.39 & 61 & 55 & 4.29 & 12.76 & 60 & 55 \\
\rowcolor{mygray}
RecBERT+PASTS (Ours) & 3.13 & 11.20 & 72 & 67 & \textbf{3.99} & 11.70 & \textbf{63} & \textbf{58} & \textbf{4.07} & 12.08 & \textbf{64} & \textbf{59} \\ \hline
HAMT \cite{chen2021history}& 2.51 & 11.15 & 76 & 72 & 2.29 & 11.46 & 66 & 61 & 3.93 & 12.27 & 65 & 60 \\
\rowcolor{mygray}
HAMT+PASTS (Ours) & \textbf{2.46} & 11.33 & \textbf{77} & \textbf{74} & 3.37 & 11.77 & \textbf{68} & \textbf{62} & \textbf{3.77} & 12.91 & \textbf{67} & \textbf{61} \\ \hline
DUET \cite{chen2022think}& 2.25 & 11.82 & 80 & 75 & 3.67 & 12.96 & 72 & 60 & 3.65 & 14.73 & 69 & 59 \\
\rowcolor{mygray}
DUET+PASTS (Ours) & \textbf{1.89} & 11.31 & \textbf{82} & \textbf{79} & \textbf{3.01} & 13.48 & 72 & \textbf{62} & \textbf{3.55} & 14.39 & \textbf{70} & \textbf{60} \\ \bottomrule
\end{tabular}}
\end{table*}
% =============
% Experiments
% =============
\section{Experiments}
\label{sub_experiments}

\subsection{Dataset}
\label{subsec_dataset}
Following previous VLN methods~\cite{tan2019learning,hong2021vln,chen2021history,chen2022think}, the standard VLN benchmark R2R dataset~\cite{anderson2018vision} is used to evaluate the performance of the PASTS model for the VLN task. The R2R dataset contains images from 90 different buildings, 21,576 navigation instructions with an average length of 29 words, and 7,189 paths with an average length of 10 meters. The R2R dataset is divided into a training set, a validation-seen set (Val Seen), a validation-unseen set (Val Unseen), and a test unseen set. The results of the test set are reported through an online challenge leaderboard. The training set and the validation-seen set cover the same 61 scenes, with the corresponding 15,045 instructions split into sets of 14,025 and 1,020, respectively. The validation-unseen set consists of 11 scenes and 2,349 instructions, and the remaining 18 scenes belong to the test set. The PREVELANT dataset~\cite{hao2020towards} is employed to provide 665,206 sampled trajectories during the back translation stage for training the follower. In this paper, the FGR2R dataset~\cite{hong2020sub} is utilized to provide annotations of sub-pairs for supervising SPM. Notably, FGR2R only includes segmentation information, rather than additional trajectory-instruction pairs other than R2R.

\subsection{Metrics}
\label{subsec_metrices}
For evaluating the instruction generation, the four typical techniques are used to evaluate the speaker: BLEU~\cite{papineni2002bleu}, ROUGE~\cite{lin2004rouge}, CIDEr~\cite{vedantam2015cider} and SPICE~\cite{anderson2016spice}. Concretely, BLEU and ROUGE focus on the accuracy and recall of the predicted sentences, respectively. CIDEr and SPICE are mainly applied in the field of image captioning. For evaluating the navigation performance, four standard metrics are used: 1) The navigation error (NE) measures the distance between the last location in prediction and in reference. 2) The success rate (SR) shows the frequency of the correct stop within a certain threshold distance of ground truth. 3) The trajectory length (TL) shows the average length of navigation. 4) The success rate weighted by the path length (SPL) considers both SR and the TL and takes both effectiveness and efficiency into account.

\subsection{Implementation Details}
\label{subsec_implementationdetails}
PASTS is trained based on the corpus from the R2R and the FGR2R dataset, where the latter is used to provide the alignment supervision signal for training SPM. The code is built based on Python and experiments are conducted on a single NVIDIA GeForce RTX 2060 GPU with Intel Core i7-9750H CPU. As in~\cite{ghaderi2022visualized}, the detailed experiment settings of PASTS are listed in Tab.~\ref{tab_experiment_settings}. The weight initialization follows the default settings in PyTorch. The model with the highest BLEU-4 on the unseen set is saved. To evaluate the impact of visual representations on the model, three widely used image feature extractors: ResNet-152~\cite{he2016deep}, ViT-B/16~\cite{dosovitskiy2020image}, and CLIP-B/32~\cite{radford2021learning} are analyzed. The dimensions of image features are 2048, 768, and 512, respectively. When training the different follower models based on back translation, the features and training strategy for the follower and PASTS keep the same as those methods to ensure a fair comparison.

\subsection{Main Results for Different Speaker Models}
\label{subsec_mainresults}
Tab.~\ref{table_resultsTIG} compares the performance of the different speaker models in generating trajectory instructions on the R2R benchmark. Speaker-Follower~\cite{fried2018speaker} uses an LSTM architecture with panoramic views as inputs. WIP-Speaker~\cite{agarwal2019visual} proposes a work-in-progress speaker model that adopts the hard attention mechanism in two stages. Env-Speaker~\cite{tan2019learning} stacks two BLSTM encoders to enhance the visual representations. Imp-Speaker~\cite{wu2021improved} and Cms-Speaker~\cite{magassouba2021crossmap} are two recent efforts to develop the speaker and the agent both based on the transformer.

The results presented in Tab.~\ref{table_resultsTIG} indicate that PASTS outperforms previous speaker models when evaluated under the same visual features, thereby achieving a new state-of-the-art performance for the VLN task. The substantial improvement in BLEU and SPICE scores suggests that PASTS generates more natural, fluent, and semantically rich trajectory descriptions compared to the existing speaker models. Notably, the observed performance gains are consistent across both the seen and unseen validation sets, indicating that PASTS exhibits strong generalization and robustness capabilities. Furthermore, when using alternative feature extractors, ViT-B/16 demonstrates superior performance compared to the other two extractors (\textit{e.g.}, SPICE 22.9 \textit{vs.} 20.9, and CIDEr 16.8 \textit{vs.} 15.2). This suggests that ViT-B/16 is more effective in capturing the visual and semantic information of the images and can potentially improve the quality of captioning.

% ============
% Ablation studies
% ============
\begin{table*}
\caption{Ablation studies of the PASTS. The values in bold represent the best results achieved in the corresponding module.}
\centering
\resizebox{\textwidth}{!}{
    \begin{tabular}{l|ccccc|ccccc}
    \toprule
    \multicolumn{1}{l|}{\multirow{2}{*}{Setting}} & \multicolumn{5}{c|}{R2R Validation-Seen} & \multicolumn{5}{c}{R2R Validation-Unseen} \\
    \multicolumn{1}{c|}{} & BLEU-1$\uparrow$ & BLEU-4$\uparrow$ & ROUGE-L$\uparrow$ & CIDEr$\uparrow$ & SPICE$\uparrow$ & BLEU-1$\uparrow$ & BLEU-4$\uparrow$ & ROUGE-L$\uparrow$ & CIDEr$\uparrow$ & SPICE$\uparrow$ \\ \midrule
    \textbf{Spatial Fusion Method.} & \multicolumn{1}{l}{} & \multicolumn{1}{l}{} & \multicolumn{1}{l}{} & \multicolumn{1}{l}{} & \multicolumn{1}{l|}{} & \multicolumn{1}{l}{} & \multicolumn{1}{l}{} & \multicolumn{1}{l}{} & \multicolumn{1}{l}{} & \multicolumn{1}{l}{} \\
    w/o spatial encoding & 56.8 & 18.2 & 36.4 & 18.0 & 22.6 & 55.7 & 16.7 & 35.7 & 13.6 & 20.5 \\
    w/ SA & 57.0 & 18.0 & 36.6 & 16.4 & 22.5 & 56.2 & 17.4 & 35.5 & 14.8 & 20.4 \\
    w/ BLSTM + SA & 59.1 & 19.2 & 37.5 & 19.1 & 24.7 & 55.7 & 17.0 & 35.5 & 14.6 & 21.4 \\
    w/ MCA & \textbf{59.6} & \textbf{19.7} & \textbf{37.8} & \textbf{19.5} & 25.6 & \textbf{57.1} & \textbf{17.6} & \textbf{36.6} & \textbf{15.2} & \textbf{21.5} \\
    w/ BLSTM + MCA & \textbf{59.6} & 19.3 & 37.5 & 19.4 & \textbf{25.7} & 56.7 & 17.4 & 36.0 & 14.6 & \textbf{21.5} \\ \midrule
    \textbf{Speaker Progress Monitor.} & \multicolumn{1}{l}{} & \multicolumn{1}{l}{} & \multicolumn{1}{l}{} & \multicolumn{1}{l}{} & \multicolumn{1}{l|}{} & \multicolumn{1}{l}{} & \multicolumn{1}{l}{} & \multicolumn{1}{l}{} & \multicolumn{1}{l}{} & \multicolumn{1}{l}{} \\
    w/o SPM & 58.4 & 18.9 & 37.4 & 18.7 & 24.4 & 56.1 & 17.2 & 36.4 & 14.9 & 21.1 \\
    w/ SPM ($\lambda=0.4$) & 59.3 & \textbf{19.8} & 37.7 & 19.0 & 24.8 & 56.4 & 16.7 & 35.9 & 14.7 & 20.4 \\
    w/ SPM ($\lambda=0.6$) & 59.4 & 19.2 & 37.4 & 19.2 & 24.5 & 56.2 & \textbf{17.6} & 36.0 & \textbf{15.3} & 21.2 \\
    w/ SPM ($\lambda=0.8$) & \textbf{59.6} & 19.7 & \textbf{37.8} & \textbf{19.4} & \textbf{25.6} & \textbf{57.1} & \textbf{17.6} & \textbf{36.6} & 15.2 & \textbf{21.5} \\ \midrule
    \textbf{Multifeature Dropout.} & \multicolumn{1}{l}{} & \multicolumn{1}{l}{} & \multicolumn{1}{l}{} & \multicolumn{1}{l}{} & \multicolumn{1}{l|}{} & \multicolumn{1}{l}{} & \multicolumn{1}{l}{} & \multicolumn{1}{l}{} & \multicolumn{1}{l}{} & \multicolumn{1}{l}{} \\
    w/o MFD & 57.8 & 18.5 & 36.2 & 16.7 & 24.1 & 55.4 & 16.2 & 35.5 & 13.6 & 20.5 \\
    w/ $\text{MFD}_{2-3}$  & 58.9 & \textbf{19.9} & 37.7 & \textbf{20.8} & 25.5 & 55.7 & 17.0 & 35.9 & 14.8 & 20.9 \\
    w/ $\text{MFD}_{1-3}$  & 59.4 & \textbf{19.9} & 37.6 & 20.4 & 25.5 & 56.2 & 17.2 & 35.9 & \textbf{15.2} & 20.3 \\
    w/ $\text{MFD}_{1-5}$  & \textbf{59.6} & 19.7 & \textbf{37.8} & 19.4 & \textbf{25.6} & \textbf{57.1} & \textbf{17.6} & \textbf{36.6} & \textbf{15.2} & \textbf{21.5} 
    \\ \bottomrule
    \end{tabular}}
\label{tab_unify_ablation}
\end{table*}

\subsection{Results in Combination with VLN Followers}
\label{subsec_result_VLNagent}
To verify the hypothesis that a more accurate speaker leads to better performance of the VLN follower by providing more precise pseudo-instructions, the trained PASTS is integrated with four recent VLN methods based on back translation. Specifically, EnvDrop~\cite{tan2019learning} uses the environmental dropout with a BLSTM structure. RecBERT~\cite{hong2021vln} designs a recurrent BERT structure for action prediction. HAMT~\cite{chen2021history} leverages a history-aware multimodal transformer to memorize past information. DUET~\cite{chen2022think} introduces a dual-scale graph transformer for long-term action planning. The first two are based on ResNet-152 features and the second two are based on ViT-B/16 features.

As shown in Tab.~\ref{table_resultsVLN}, PASTS achieves significant improvements compared with the existing methods. Concretely, the improvement for EnvDrop is the most obvious, where SR is improved by 7\%, 3\%, and 5\% and SPL is improved by 6\%, 2\%, and 5\% on the validation-seen, validation-unseen and test-unseen sets, respectively. The improvements for the other three models are smaller since they have been pre-trained in the first stage. Nevertheless, the experimental results still prove that PASTS can further improve the robustness of these pre-trained VLN models. For instance, with the pseudo labels generated by PASTS, the state-of-the-art DUET~\cite{chen2022think} achieves 4\% SPL and 2\% SR improvement on the validation-seen set. This indicates that the proposed approach allows the generative pseudo instructions to align better with the sampled trajectories, thereby reducing the potential risks of noise. These findings suggest that PASTS is a model-agnostic approach that can effectively enhance the performance of existing VLN models. Its ease of implementation and potential for improving learning outcomes make it a promising tool for further research in the field.

\subsection{Ablation Studies}
\label{subsec_ablation study}

In this section, ablation studies based on ResNet-152 are presented to validate the contributions of the proposed modules to the effectiveness of the PASTS framework. The results are reported in Tab.~\ref{tab_unify_ablation}.

\subsubsection{Effect of the different spatial fusion methods} 
In Tab.~\ref{tab_unify_ablation}, ``w/o spatial encoding" means that only action features are used. ``SA" and ``MCA" represent the soft attention mechanism and the multihead cross attention mechanism, respectively. ``BLSTM" denotes the bidirectional LSTM encoding applied to the action features. The results demonstrate that the fusion of panoramic information can effectively improve the richness of effective information, but different fusion techniques will produce varied results. It shows that the spatial encoder with mutlihead cross-modal attention increases the performance on all metrics(\textit{e.g.}, BELU-4 19.7 \textit{vs.} 18.5, and SPICE 25.6 \textit{vs.} 23.9). The multihead structure enables the model to attend to various parts of the inputs and learn more semantically rich visual representations, enhancing the ability to observe and comprehend its surroundings. Moreover, the addition of BLSTM to action features does not significantly improve performance and may even decrease certain metrics in the unseen environment. To avoid introducing unnecessary parameters, a spatial encoder is designed using cross-modal multihead attention to capture visual observations in space.

\subsubsection{Effect of the SPM} 
A joint loss function with two hyperparameters is proposed to train the model in an end-to-end manner. The parameter $\lambda$ balances the SWP and SPM losses, and $\omega$ ensures they are of the same magnitude. Based on experiments, we have found that the loss of the SWP is approximately 10 times the loss of the SPM. Therefore, to unify the magnitudes of these two losses, we set $\omega=10$. As shown in Tab.~\ref{tab_unify_ablation}, the model with $\lambda=0.8$ achieves the best performance. This configuration results in significant improvements across all metrics, particularly with a notable increase in SPICE from 24.4 to 25.6 on the seen validation split. It is worth noting that when the weight of the SPM surpasses that of the original loss, such as with $\lambda=0.4$, the model's performance decreases. This outcome is attributed to the priority of the auxiliary task, which should not outweigh that of the main task. Otherwise, the model may focus more on optimizing the auxiliary task, which can lead to suboptimal performance. As a result, the proposed SPM is effective in improving the alignment and coherence of the generated instructions with the navigation stages, which enables PASTS to better fit the trajectory and produce instructions with greater fluency and richer semantic content.
% ==========
% Modules on VLN follower
% ==========
\begin{table}[tb]
\caption{Performances of different modules on follower training.}
\centering
\label{tab_ablation_agent}
\renewcommand{\arraystretch}{1.1}
\begin{tabular}{c|ccc|cc|cc}
\toprule
 \multirow{2}{*}{Method} & \multirow{2}{*}{ST}  & \multirow{2}{*}{SPM} & \multirow{2}{*}{MFD} & \multicolumn{2}{c|}{Val Seen} & \multicolumn{2}{c}{Val Unseen} \\  
 & & & & SR$\uparrow$ & SPL$\uparrow$ & SR$\uparrow$ & SPL$\uparrow$ \\ \midrule
HAMT &  &  &  & 75.61 & 72.18 & 66.24 & 61.55 \\ \hline
+PASTS & \checkmark &  &  & 76.79 & 72.49 & 66.58 & 60.29 \\
+PASTS & \checkmark & \checkmark &  & 76.90 & 71.89 & 67.69 & 61.90 \\
\textbf{+PASTS} & \textbf{\checkmark} & \textbf{\checkmark} & \textbf{\checkmark} & \textbf{77.47} & \textbf{73.54} & \textbf{68.28} & \textbf{62.37} \\ \bottomrule
\end{tabular}
\end{table}
% ============
% instruction generation comparison
% ============
\begin{figure*}[!htb]
\includegraphics[scale=.55]{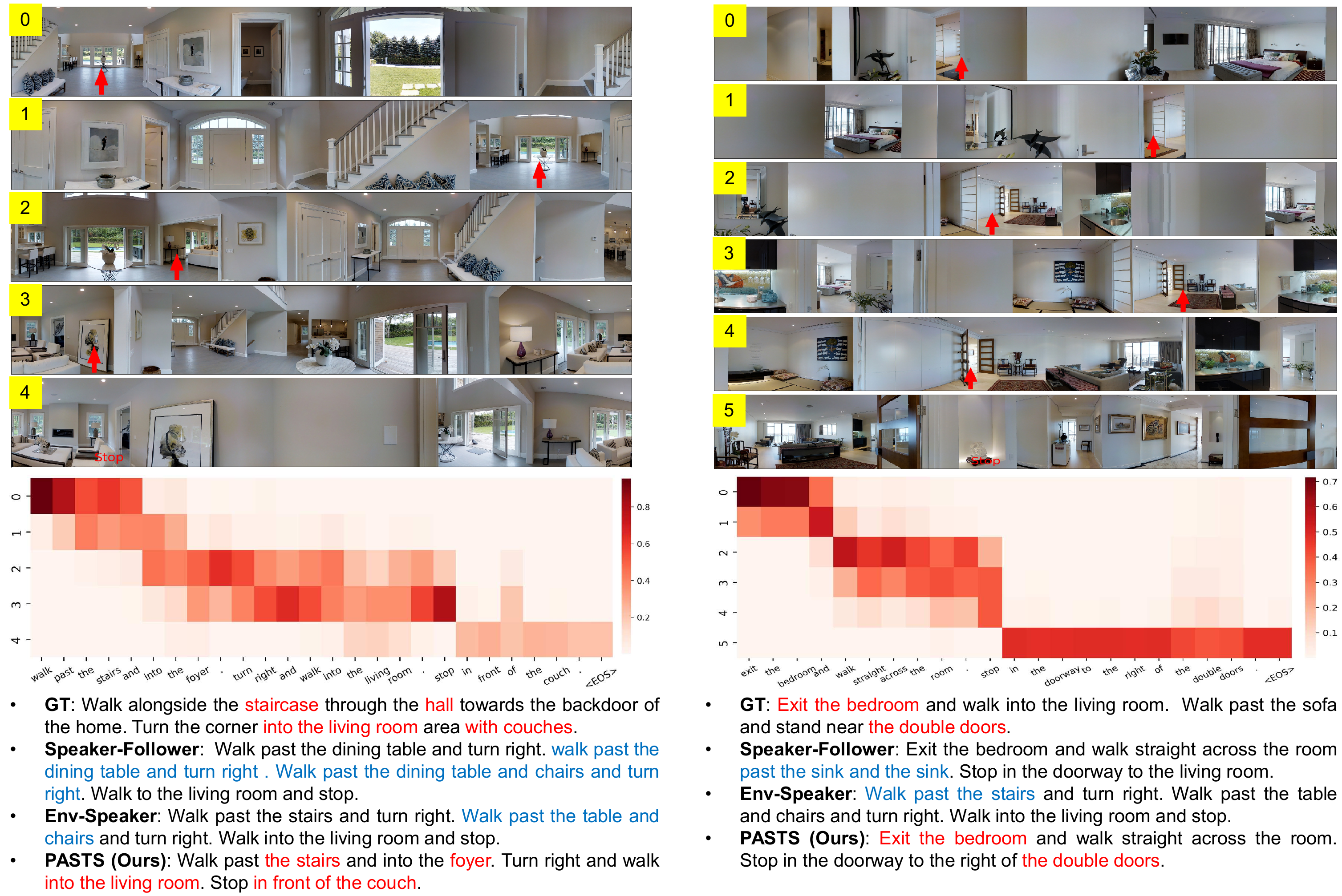}
\centering
\caption{Visualization of some generated instructions compared with the ground truth (GT) and previous methods. The top panels show panorama observations of navigation paths. The middle panels illustrate the attention values in the last decoder layer. The x and y-axis represent the predicted word and the step along the path. The bottom sentences show predictions using different methods. Colors indicate failure (blue) and outstanding predictions (red).}
\label{fig_results_heatmap}
\end{figure*}
% ============
% Learning curves
% ============
\begin{figure}
    \centering
    \includegraphics[scale=0.64]{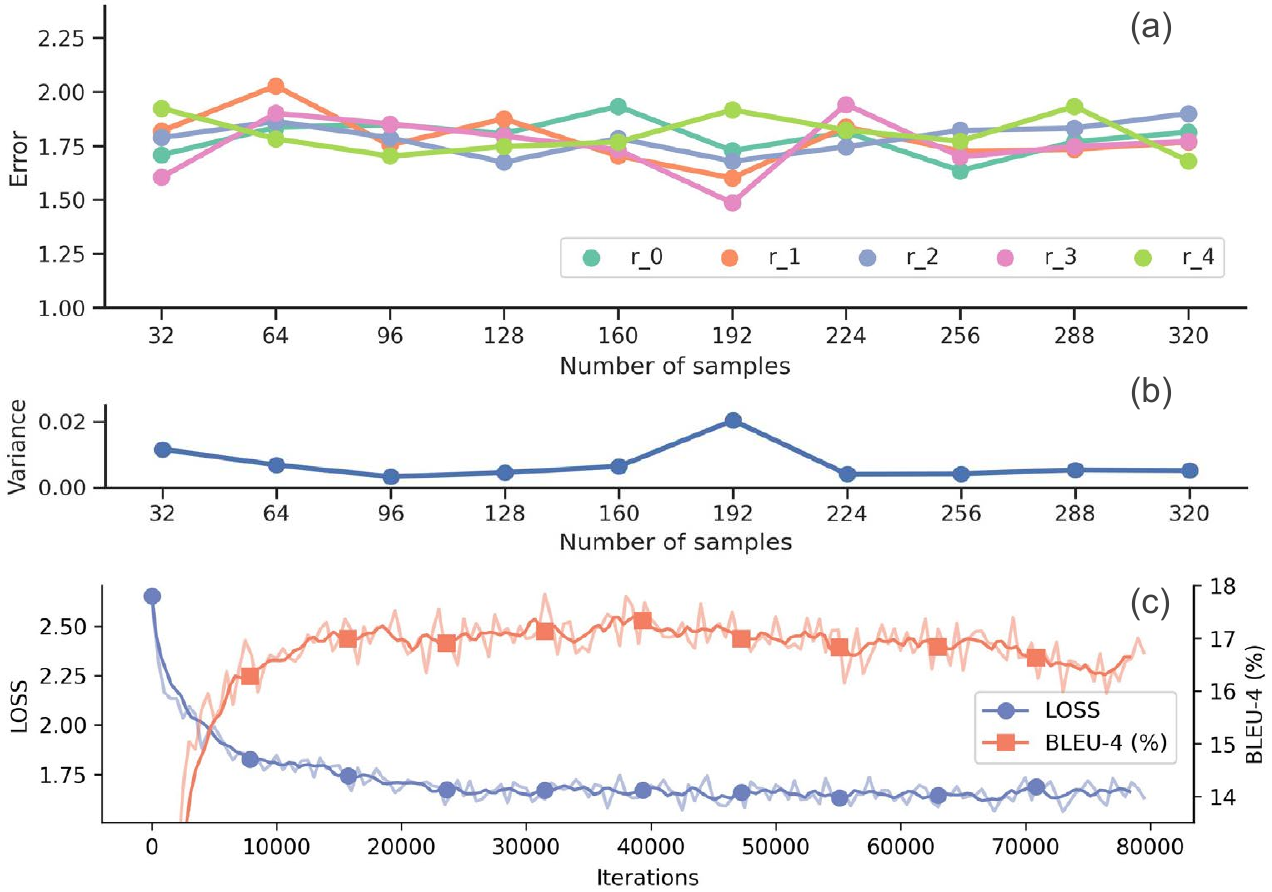}
    \caption{Analysis of uncertainty and convergence. (a) presents the values of cross entropy performed by 5 different rounds with dropout. (b) shows the variance values. (c) is the illustration of the learning curves.}
    \label{fig_learning_curve}
\end{figure}
\subsubsection{Effect of the MFD} The issue of overfitting is a significant challenge for transformer-based architectures, especially when dealing with limited and specialized datasets. To address this problem, MFD which comprises five specific types of feature dropout is suggested in this work. The results demonstrate that using the full MFD (denoted as $\text{MFD}_{1-5}$) results in significant improvements across all metrics in both the seen and unseen environment, compared with the model without MFD. Additionally, compared with the initial $\text{MFD}_{2-3}$, the join of $\text{MFD}_1$ increases BELU-1 and BLEU-4 to some extent on the unseen set, but some other metrics like CIDEr and SPICE decrease, which demonstrates that $\text{MFD}_{1-3}$ is effective but not sufficient. The further introduction of $\text{MFD}_{4-5}$ increases all metrics on the unseen set (\textit{e.g.}, BLEU-1 57.1 \textit{vs.} 56.2, and SPICE 21.5 \textit{vs.} 20.3), which proves that the addition of extra feature dropouts enables PASTS to better generalize and overcome the overfitting problem.

\subsubsection{Effect of different modules on training followers} In addition to exploring the impact of different modules on the speaker, experiments on the effects of promoting follower training are conducted. As shown in Tab.~\ref{tab_ablation_agent}, where ST denotes the spatio-temporal transformer structure, it can be seen that PASTS with the proposed modules successfully boost the navigation performance of HAMT~\cite{chen2021history} in both seen and unseen environments. For instance, incorporating SPM leads to an improvement in SR on the unseen set from 66.58 to 67.69. Moreover, with the inclusion of MFD, there are larger improvements in SR and SPL on the unseen split, from 66.24 to 68.28, and from 61.55 to 62.37, respectively. This finding suggests that a more robust speaker has the potential to provide more precise pseudo instructions, thereby minimizing the interference that could potentially impede the follower training process. 

% ==========
% Visualize_SPM
% ==========
\begin{figure}[tb]
\includegraphics[scale=.6]{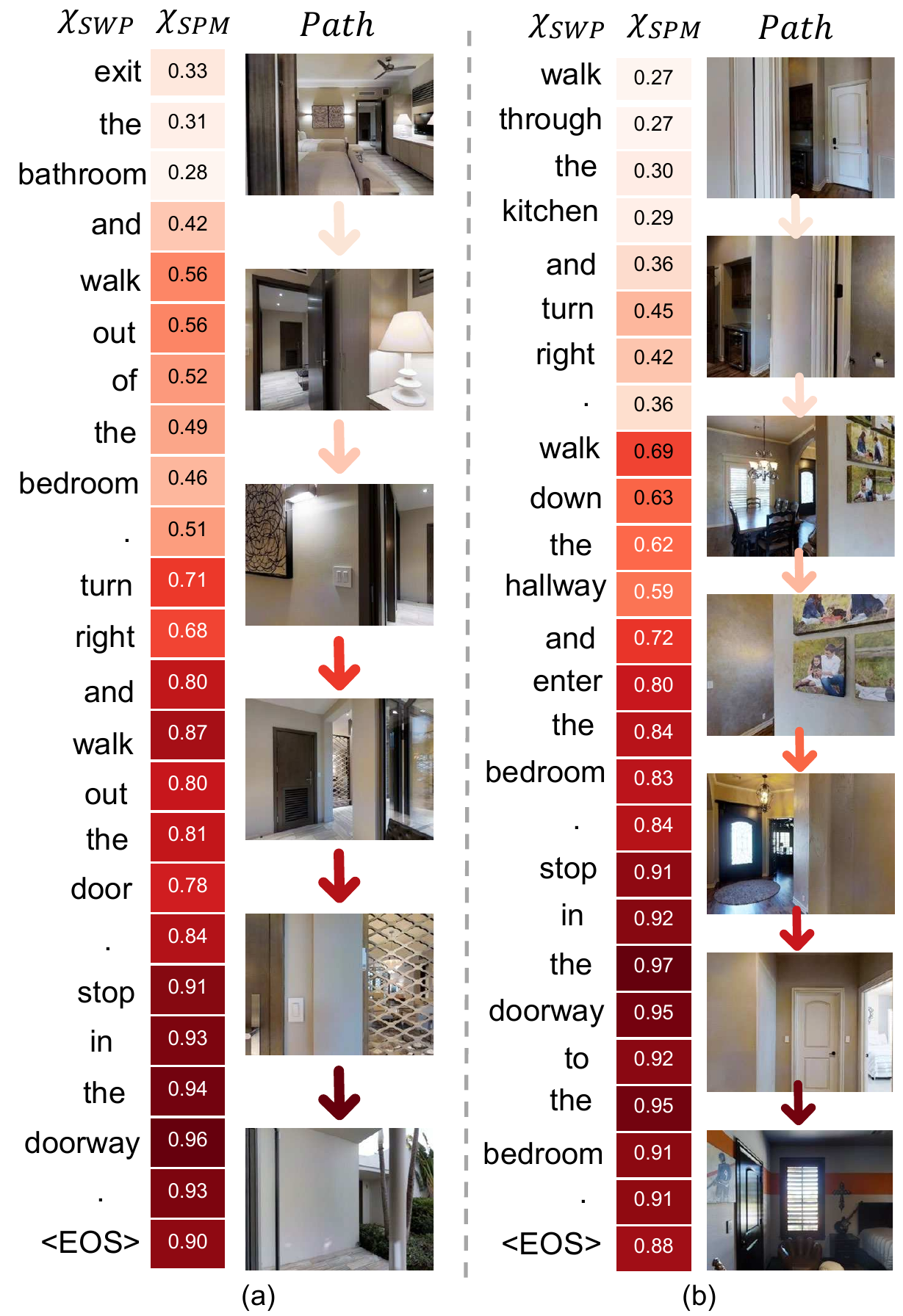}
\centering
\caption{Illustration of the predicted words $\mathcal{X}_{SWP}$ with the predicted progress $\mathcal{X}_{SPM}$ output by PASTS. The corresponding images of paths are given in the right.}
\label{fig_visual_spm}
\end{figure}
\begin{figure*}[tbp]
\includegraphics[scale=.55]{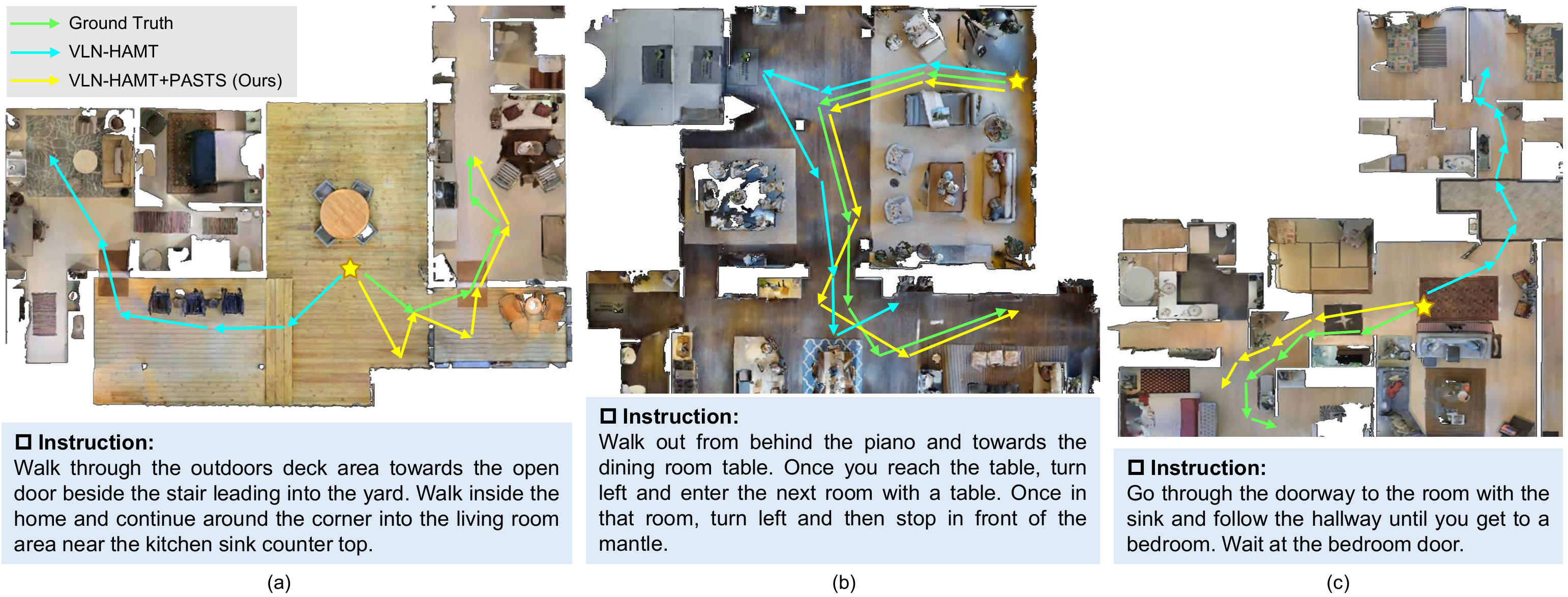}
\centering
\caption{Comparison of navigation results obtained using different approaches. The star icon denotes the starting point, and arrows in different colors represent different navigation results. The instructions for guidance are shown in the bottom.}
\label{fig_visual_navigation}
\end{figure*}

\subsection{Visualization}
\label{subsec_visualization}
\subsubsection{Heatmap of Vision-and-Language Decoder} To further validate the proposed method compared with previous methods, visualized examples are shown in Fig.~\ref{fig_results_heatmap}. It can be seen that each predicted word corresponds well to the navigation phase. For example, when the model outputs ``walk past the stairs and into the foyer", it focuses on the first three viewpoints, where there is indeed a stair and a hall in the pictures. Additionally, PASTS can better capture the dominant object and meanwhile reduce the repeated generation issue. For example, in the left case, Speaker-Follower outputs ``walk past the dining table and turn right" many times, and Env-Speaker ignores the key landmark ``hall". As a result, compared with previous methods, PASTS successfully mitigates the repetition problem and enhances the richness of the generated visual representations.

\subsubsection{Analysis of Uncertainty and Convergence}
The uncertainty analysis is conducted on the optimum model using the Monte-Carlo Dropout (MCD) method~\cite{gal2016dropout,abbaszadeh2022novel}, and the results are presented in Fig.~\ref{fig_learning_curve} (a) and (b). The analysis involves running the model with dropout enabled for 5 rounds and measuring the variance of the results, which is found to be 0.0071. Fig.~\ref{fig_learning_curve} (c) shows the learning curves of the loss and BLEU-4 on the unseen set. The results demonstrate that the model exhibits good convergence and stability, with the uncertainty variance being controlled within a small range. Additionally, the model is capable of quickly reaching the training peak, indicating efficient convergence.

\subsubsection{Speaker Progress Prediction}
The progress values output by the SPM prediction head and the predicted words are visualized in Fig.~\ref{fig_visual_spm}. For each example, the left to right columns denote the predicted words, predicted progress, and the given paths, respectively. The heatmaps of progresses show that the gaps between different phrases with respect to different navigation stages are obvious. For example, in (a), the average progress value of ``exit the bathroom'' is about 0.31, and that of ``walk out of the bedroom'' is about 0.52. This demonstrates PASTS is capable of recognizing the correlation between phrases and paths, which can assist the speaker in producing more fine-grained instructions for lengthy trajectories.

\subsubsection{Navigation Results} As described in Sec.~\ref{sub_training}, PASTS is employed to provide more sorts of pseudo instructions based on the back translation method to enhance the generalization of initial follower models. Some visualized instances of predicted pathways on the validation set are displayed in Fig.~\ref{fig_visual_navigation}. It demonstrates that the generalization of the HAMT model significantly improves with the aid of the data augmentation given by PASTS. The trajectories predicted by the follower trained with PASTS are more consistent with the given natural instructions and can get close to the intended destination. However, the model without PASTS augmentation may lead to the wrong ending location (as shown in (b)) or completely wrong path predictions (as shown in (a) and (c)). This is due to the limited instructions and environments from the original dataset during training, which could lead to the misinterpretation of the unseen inputs.

% =============
% Limitations and Future work
% =============
\section{Limitations and Future Work}
\label{sub_limitations}
\begin{figure}[!htbp]
\centering
\includegraphics[scale=0.4]{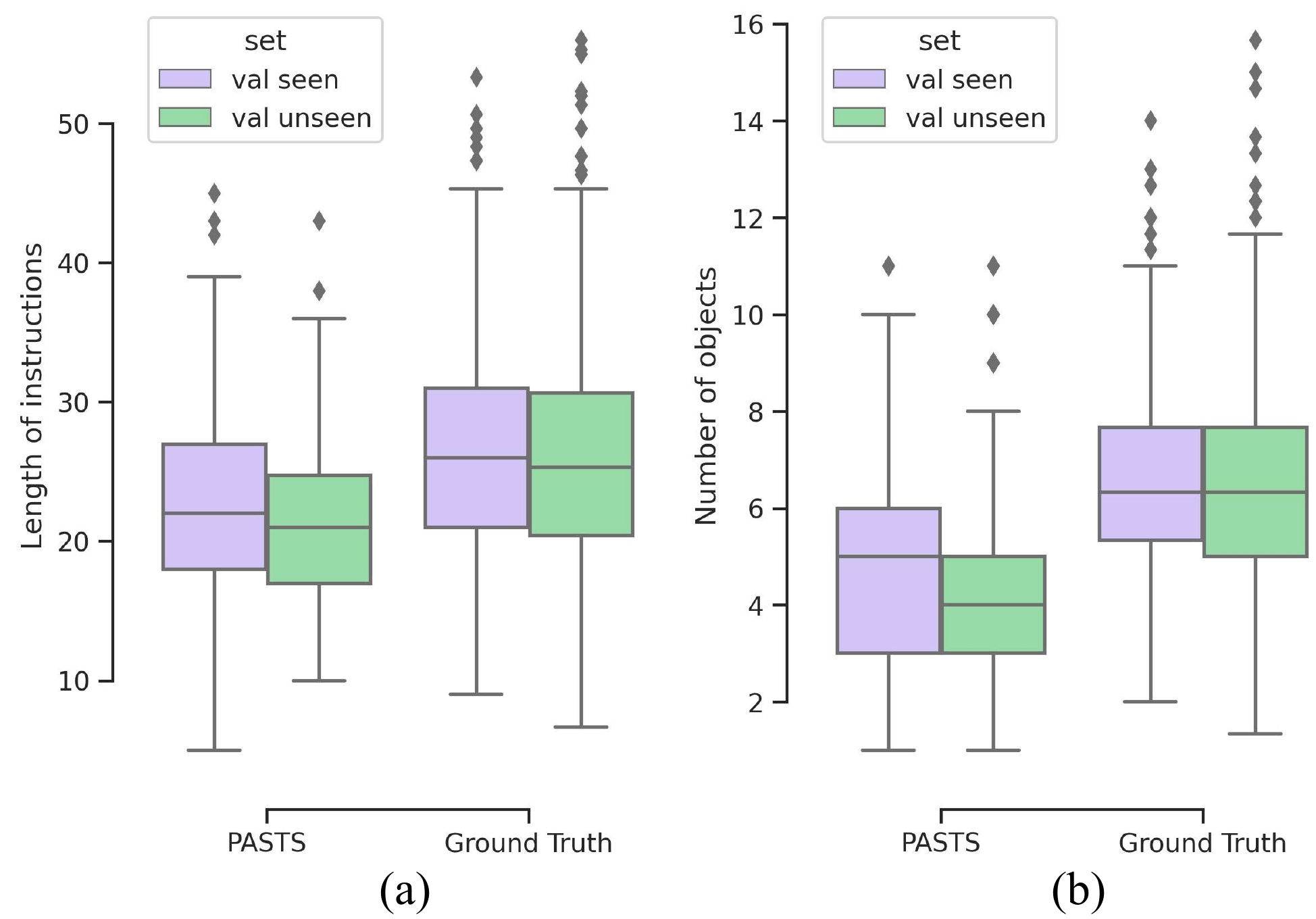}
\caption{Statistical results of data distribution. (a) shows the distribution of the length of instructions and (b) presents the distribution of the number of object words.}
\label{fig_limitation}
\end{figure}

While the proposed method has shown promising results on the VLN task, some limitations of our approach are discussed in this section to inspire future work. Fig.~\ref{fig_limitation} (a) demonstrates that the generated instructions have a biased tendency to be shorter than human-labeled instructions, which may reduce detail and richness. This could be attributed to the cross-entropy optimization used during training, which is designed to minimize the difference between the predictions and the ground truth. In practice, this may encourage the model to focus on the most salient information in the trajectory rather than some redundant descriptions that could lead to extra loss. Future work may explore reinforcement learning techniques to reward the model for generating instructions that are not only accurate but also informative and detailed. Additionally, although PASTS can capture more semantic information than previous speakers, it is still less than ground truth, as shown in Fig.~\ref{fig_limitation} (b). The possible reason could be the lack of object-level visual representations. PASTS currently relies solely on image features to generate instructions. Future work could include incorporating object detection and recognition modules into the model. Moreover, the expansion of the corpus or the analysis of different roles of parameters~\cite{asheghi2020updating} may also help to further improve the performance, which we leave for future exploration.
% =============
% Conclusion
% =============
\section{Conclusion}
\label{sub_conclusions}
In this paper, a novel progress-aware spatio-temporal transformer speaker (PASTS) is proposed to perform more accurate data augmentation for the VLN task. To better consider and fuse the sequenced action and environment information, a spatio-temporal transformer encoder is established to incorporate features both in the spatial- and temporal-dimension. The SPM module is proposed to work as an auxiliary task to settle the potential misaligned problem, and a novel joint loss function is designed to enable end-to-end training. Additionally, the MFD strategy with five types of feature dropout is explored to alleviate the overfitting problem. Based on the back translation method, PASTS can provide more fine-grained and precise pseudo instructions to effectively expand the scale of the dataset during the follower's training procedure. The experiments confirm that PASTS leads to improved performance and achieves state-of-the-art performance for both speaker and follower in VLN. This demonstrates the flexibility and robustness of PASTS, which is also expected to be helpful for other tasks related to image and video captioning.

Finally, it is worth noting that the research on the speaker itself is of great importance. The ability of an embodied agent to accurately and fluently describe a path or a series of events can greatly enhance the user's experience in human-computer interaction, and has enormous potential for future artificial intelligence research. 

\section*{Acknowledgments}
This paper is supported by the National Natural Science Foundation of China under Grants (62233013, 62073245, 62173248). Suzhou Key Industry Technological Innovation-Core Technology R\&D Program (SGC2021035); Special funds for Jiangsu Science and Technology Plan (BE2022119). Shanghai Municipal Science and Technology Major Project (2021SHZDZX0100) and the Fundamental Research Funds for the Central Universities. Shanghai Science and Technology Innovation Action Plan (22511104900).

% =============
% References
% =============
\bibliographystyle{elsarticle-num-names}
\bibliography{main}

\end{document}